\documentclass[runningheads]{llncs}
\usepackage{graphicx}

\usepackage{tikz}
\usepackage{comment}
\usepackage{amsmath,amssymb} 
\usepackage{color}
\usepackage{graphicx}
\usepackage{subfigure}

\usepackage{booktabs}
\usepackage{algorithm}
\usepackage{algorithmic}

\usepackage{amsmath}
\usepackage{amssymb}
\usepackage{amsmath,amssymb,amsfonts}
\usepackage{algorithm}
\usepackage{booktabs}
\usepackage{algorithmic}
\usepackage{bm}
\usepackage{multirow}
\usepackage{amssymb}
\usepackage{mathtools}
\usepackage{makecell}
\usepackage{enumitem}
\usepackage{bbding}
\usepackage[accsupp]{axessibility}  
\usepackage{orcidlink}

\begin{document}
\pagestyle{headings}
\mainmatter
\def\ECCVSubNumber{3827}  

\title{MixSKD: Self-Knowledge Distillation from Mixup for Image Recognition} 

\titlerunning{MixSKD: Self-Knowledge Distillation from Mixup for Image Recognition}
%
\author{Chuanguang Yang\inst{1,2}\orcidlink{0000-0001-5890-289X} \and
	Zhulin An\inst{1(}\Envelope\inst{)}\orcidlink{0000-0002-7593-8293} \and
	Helong Zhou\inst{3} \and Linhang Cai\inst{1,2} \and Xiang Zhi\inst{1,2} \and Jiwen Wu\inst{1} \and Yongjun Xu\inst{1}\orcidlink{0000-0001-6647-0986} \and Qian Zhang\inst{3}}
\authorrunning{C. Yang et al.}
%
\institute{Institute of Computing Technology, Chinese Academy of Sciences, Beijing, China \and
	University of Chinese Academy of Sciences, Beijing, China \and
	Horizon Robotics \\ 
	\email{\{yangchuanguang, anzhulin, cailinhang19g, zhixiang20g, xyj\}@ict.ac.cn, \{helong.zhou, qian01.zhang\}@horizon.ai} }

\maketitle

\begin{abstract}
	Unlike the conventional Knowledge Distillation (KD), Self-KD allows a network to learn knowledge from itself without any guidance from extra networks. This paper proposes to perform Self-KD from image Mixture (MixSKD), which integrates these two techniques into a unified framework. MixSKD mutually distills feature maps and probability distributions between the random pair of original images and their mixup images in a meaningful way. Therefore, it guides the network to learn cross-image knowledge by modelling supervisory signals from mixup images. Moreover, we construct a self-teacher network by aggregating multi-stage feature maps for providing soft labels to supervise the backbone classifier, further improving the efficacy of self-boosting. Experiments on image classification and transfer learning to object detection and semantic segmentation demonstrate that MixSKD outperforms other state-of-the-art Self-KD and data augmentation methods. The code is available at https://github.com/winycg/Self-KD-Lib.
	
\end{abstract}
\section{Introduction} 
Knowledge Distillation (KD)~\cite{hinton2015distilling} is an effective paradigm to enable a given student network to generalize better under the guidance of a pre-trained high-performance teacher. The seminal KD guides the student to mimic  the predictive class probability distributions (also namely \emph{soft labels}) generated from the teacher. Interestingly, although the soft label assigns probabilities to incorrect classes, the relative probability distribution also encodes meaningful information on similarity among various categories~\cite{yuan2020revisiting}. The soft label has been widely demonstrated as dark knowledge to enhance the student's performance~\cite{hinton2015distilling,yuan2020revisiting}.

\begin{figure}[tbp]  
	\centering 
	\includegraphics[width=0.75\linewidth]{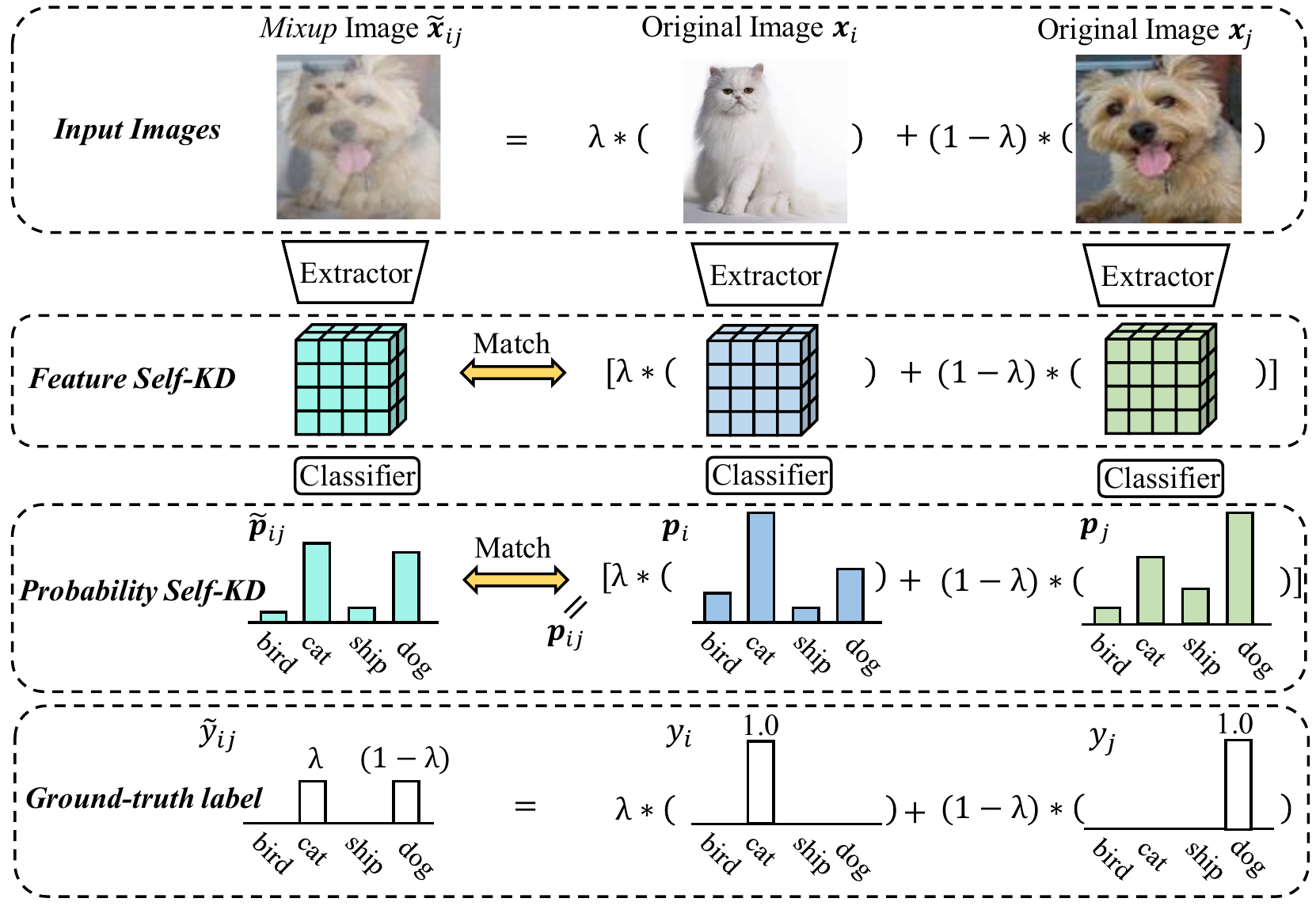}
	\caption{Overview of the basic idea about our proposed MixSKD.} 
	\label{overview_self_kd}
\end{figure}

The conventional KD~\cite{hinton2015distilling,romero2014fitnets,yang2021hierarchical,yang2022cross,yang2022knowledge} relies on a pretrained teacher to provide soft labels. Thus the two-stage process increases the training pipeline and is time-consuming. Recent studies of \emph{Self-KD} have illuminated that a network can distill its own knowledge to teach itself without a pretrained teacher. However, exploring knowledge from the network itself is a non-trivial problem. Existing Self-KD works often employ \emph{auxiliary architecture}~\cite{sun2019deeply,zhang2019your,zhang2020auxiliary,ji2021refine,zhang2021self} or \emph{data augmentation}~\cite{xu2019data,yun2020regularizing} to capture additional knowledge to enhance the network. The auxiliary-architecture-based method often appends extra branches to the backbone network. Self-KD regularizes these branches and the backbone network to generate similar predictions via knowledge transfer~\cite{sun2019deeply,zhang2019your}. Another vein is to mine meaningful knowledge from external data augmentations. Data-augmentation-based methods aim to regularize the consistent predictions between distorted versions of the same instance~\cite{xu2019data} or two instances from the same class~\cite{yun2020regularizing}. A common point of previous Self-KD methods is that the mined supervisory signal, \emph{e.g.} soft label,  is generated from an individual input sample.

In contrast, we propose incorporating \textbf{S}elf-\textbf{KD} with image \textbf{Mix}ture (namely MixSKD) into a unified framework. Image mixture has been developed as an advanced data augmentation called Mixup~\cite{zhang2017mixup}. Mixup performs pixel-wise image mixture between two randomly sampled images $\bm{x}_{i}$ and $\bm{x}_{j}$ to construct a virtual sample $\tilde{\bm{x}}_{ij}$. The virtual sample's label $\tilde{y}_{ij}$ is a linear interpolation between the one-hot ground-truth labels of $y_{i}$ and $y_{j}$ with the same mixture proportion. Additional Mixup images introduce more samples in the input space by randomly fusing different images. This allows Self-KD to distill richer knowledge towards image mixture over the independent samples. The success behind Mixup is to encourage the network to favour simple linear behaviours in-between training samples~\cite{zhang2017mixup}. Our MixSKD further takes advantage of this property by modelling supervisory signals from Mixup images in the feature and probability space.

As shown in Fig.~\ref{overview_self_kd}, we propose to linearly interpolate the probability distributions $\bm{p}_{i}$ and $\bm{p}_{j}$ inferred from the original images $\bm{x}_{i}$ and $\bm{x}_{j}$ to model the soft ensemble label $\bm{p}_{ij}$ to supervise the Mixup distribution $\tilde{\bm{p}}_{ij}$. Intuitively, $\bm{p}_{ij}$ encodes the crude predictions based on the full information in two original images. It can be seen
as a pseudo teacher distribution to provide comprehensive
knowledge for Mixup-based $\tilde{\bm{p}}_{ij}$. Symmetrically, we also regard $\tilde{\bm{p}}_{ij}$ as the soft label to supervise $\bm{p}_{ij}$. $\tilde{\bm{p}}_{ij}$ could be regarded as a data-augmented distribution to refine $\bm{p}_{ij}$ for learning robust mixed predictions under Mixup and avoid overfitting. The mutual distillation process encourages the network to produce consistent predictions between the pair of original images $(\bm{x}_{i},\bm{x}_{j})$ and their Mixup image $\tilde{\bm{x}}_{ij}$. Another efficacy compared with the conventional Mixup training is that Self-KD would force the network to generate similar wrong predictions over incorrect classes between $\bm{p}_{ij}$ and $\tilde{\bm{p}}_{ij}$. This property of dark knowledge may also be a critical factor demonstrated by previous KD research~\cite{yuan2020revisiting}.

Inspired by~\cite{zhang2019your,sun2019deeply,yang2021hierarchical,yang2022knowledge}, we append auxiliary feature alignment modules for transforming the feature maps from shallow layers to match the hierarchy with the final feature map. We match these feature maps between the pair of original images $(\bm{x}_{i},\bm{x}_{j})$ and their Mixup image $\tilde{\bm{x}}_{ij}$ via Self-KD, as shown in Fig.~\ref{overview_self_kd}. Motivated by~\cite{wang2018adversarial}, we propose an adversarial feature Self-KD method that utilizes $l_{2}$ loss to force the feature maps to be close and a discriminator loss to simultaneously increase the difficulty of mimicry. This process encourages the network to learn common semantics between matched feature maps. We further attach auxiliary classifiers to these hidden feature maps to output probability distributions and perform probability Self-KD. This further regularizes the consistency of intermediate information.

Over the training graph, we also construct a self-teacher to supervise the final classifier of the backbone network. The self-teacher aggregates all intermediate feature maps and uses a linear classifier to provide meaningful soft labels. The soft label is informative since it assembles all feature information across the network. During the training phase, we introduce auxiliary branches and a self-teacher to assist Self-KD.  During the test phase, we discard all auxiliary architectures, resulting in no extra costs compared with the baseline. 

Following the consistent benchmark~\cite{yun2020regularizing,ji2021refine}, we demonstrate that MixSKD is superior to State-Of-The-Art (SOTA) Self-KD methods and data augmentation approaches on image classification tasks. Extensive experiments on downstream object detection and semantic segmentation further show the superiority of MixSKD in generating better feature representations.

The contributions are three-fold: (1) We propose incorporating Self-KD with image mixture into a unified framework in a meaningful manner to improve image recognition. (2) We construct a self-teacher by aggregating multi-stage feature maps to produce high-quality soft labels. (3) MixSKD achieves SOTA performance on image classification and downstream dense prediction tasks against other competitors.

\section{Related Works}
\textbf{Multi-Network Knowledge Distillation.} The conventional multi-network KD often depends on extra networks for auxiliary training in an offline or online manner. The offline KD~\cite{hinton2015distilling,ge2019distilling,peng2019correlation,yang2021hierarchical,yang2022multi,yang2022cross,yang2022knowledge,liu2022coupleface,liu2022C3} transfers knowledge from a pre-trained teacher to a smaller student. Online KD~\cite{zhang2018deep,zhu2018knowledge,chen2020online,yang2021multi,yang2022mutual,yang2022focal,yang2022knowledge} aims to train multiple student networks and perform knowledge transfer within the cohort in an online manner. A popular mechanism called \emph{Deep Mutual Learning} (DML)~\cite{zhang2018deep} or \emph{Mutual Contrastive Learning} (MCL)~\cite{yang2022mutual} suggests that an ensemble of students learn significantly better through teaching each other. Our MixSKD also employs mutual mimicry of intermediate feature maps and probability distributions. An important distinction with DML is that we aim to distill information from the view of Mixup \emph{within a single network}. In contrast to mutual learning, another vein is to construct a virtual online teacher via knowledge ensembling from multiple student networks~\cite{zhu2018knowledge}. Instead of multiple networks, we aggregate multi-stage feature maps from attached branches \emph{within a single network} to construct a powerful self-teacher. This self-teacher generates excellent soft labels for auxiliary Mixup training.

\textbf{Self-Knowledge Distillation.} Unlike offline and online KD, Self-KD aims to distill knowledge from the network itself to improve its own performance. Self-KD is a non-trivial problem since it does not have explicit peer networks. Therefore, a natural idea is to introduce \emph{auxiliary architecture} to capture extra knowledge. DKS~\cite{sun2019deeply} proposes pairwise knowledge alignments among auxiliary branches and the primary backbone. BYOT~\cite{zhang2019your} regards the deepest classifier as a teacher and transfers knowledge to shallow ones. ONE~\cite{zhu2018knowledge} aggregates an ensemble logit from multiple auxiliary networks to provide the soft label. Beyond the logit level, SAD~\cite{hou2019learning} performs top-down and layer-wise attention distillation within the network itself. FRSKD~\cite{ji2021refine} generates a self-teacher network for itself to provide refined feature maps for feature distillation. Another orthogonal aspect is to excavate knowledge from \emph{data augmentation}. DDGSD~\cite{xu2019data} and CS-KD~\cite{yun2020regularizing} regularize the consistency of predictive distributions between two different views. DDGSD utilizes two different augmented versions of the same image as two views, while CS-KD leverages two different samples from the same class. Moreover, prediction penalization is also a form of Self-KD. Tf-KD~\cite{yuan2020revisiting} connects label smoothing~\cite{szegedy2016rethinking} to Self-KD and applies a manually designed soft label to replace the teacher. Orthogonal to the above studies, we propose Self-KD from the image mixture perspective and achieve the best performance. 

\textbf{Image Mixture.} Image mixture has been developed to a robust data augmentation strategy. The seminal Mixup~\cite{zhang2017mixup} performs global pixel-wise interpolations between two images associated with the same linear interpolation of one-hot labels. Wang \emph{et al.}~\cite{wang2020neural} employ Mixup to augment a few unlabeled images for data-efficient KD. Beyond input space, Manifold Mixup~\cite{verma2019manifold} conducts interpolations in the hidden feature representations. The success behind image mixture is to provide mixed training signals for regularizing the network to behave linearly. \emph{However, none of the previous work explores Mixup training signals as extra knowledge to model soft labels for Self-KD.}  Our MixSKD incorporates Mixup with Self-KD into a unified framework and regularizes the network more meaningfully. Although previous CS-KD and FRSKD also attempted to combine Mixup to further enhance the performance, they often utilized a straightforward instead of a thoughtful way. Yun \emph{et al.}\cite{yun2019cutmix} observed that Mixup samples are locally unnatural, confusing the network for object localization. CutMix~\cite{yun2019cutmix} is proposed to execute patch-wise mixture between two images. We remark that our MixSKD can also combine CutMix to regularize the spatial consistency between patch-wise information mixture. This may become a promising future work.

\begin{figure}[tbp]  
	\centering 
	\includegraphics[width=1.\linewidth]{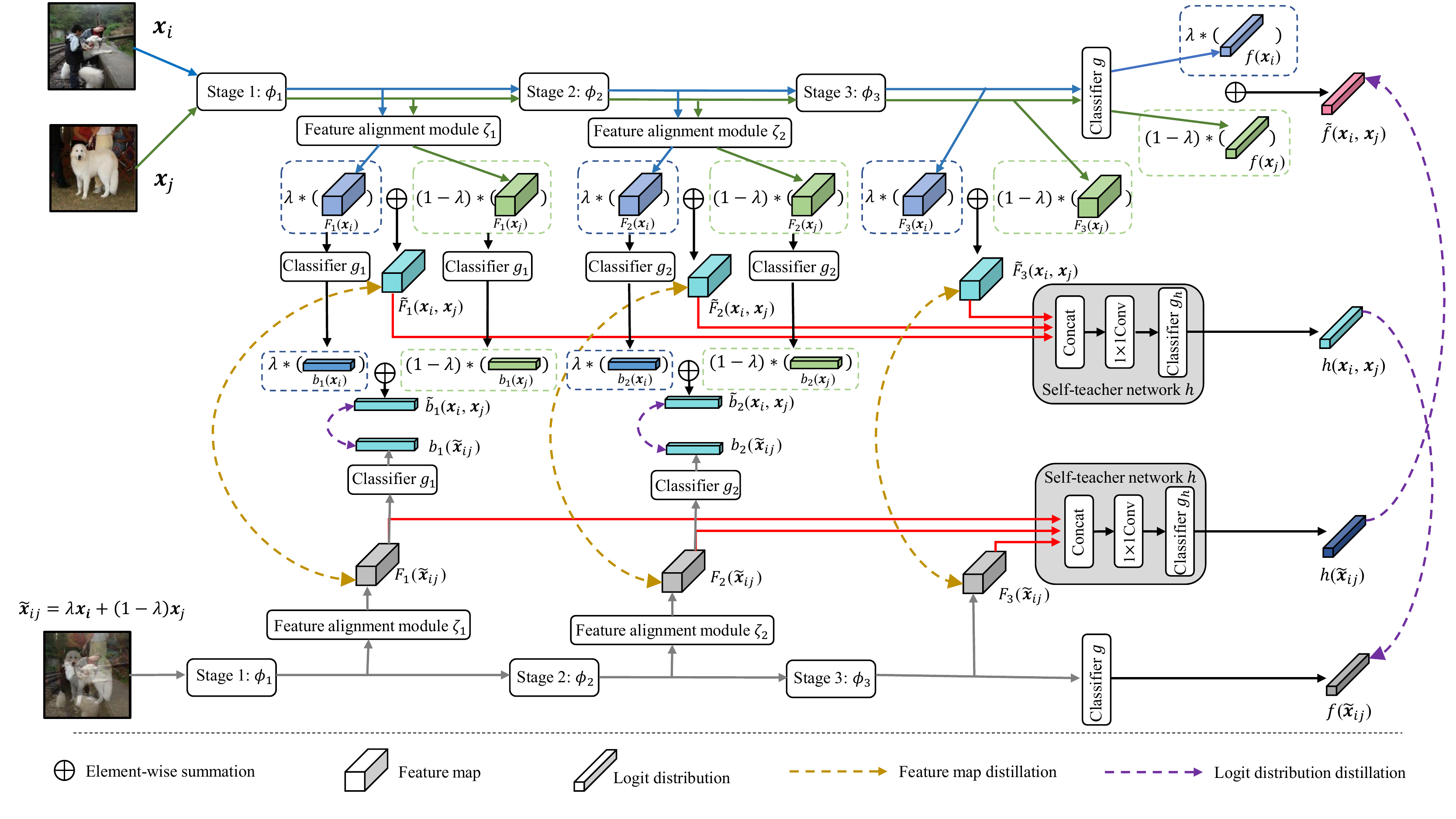}
	\caption{Overview of the proposed MixSKD over a network with $K=3$ stages. We employ a shared architecture between the training graph of the original input images (\emph{upper}) and that of the Mixup input images (\emph{lower}). The meta-architectures with the same notation denote the identical one. During the training stage, we regularize the \emph{feature maps} and \emph{class probability distributions} between the original images and Mixup images via distillation losses. During the test stage, we drop all auxiliary components.} 
	\label{MixSKD}
\end{figure}

\section{Methodology}

\subsection{Formulation of Training Graph} The convolutional neural network (CNN) $f$ for image classification can be composed of a feature extractor $\phi $ and  a linear classifier $g$, \emph{i.e.} $f= g\circ \phi $. For simplicity, we omit \emph{Global Average Pool} between feature extractor and linear classifier. The feature extractor $\phi $ often contains multiple stages for refining feature hierarchies that is formulated as $\phi=\phi_{K}\circ \cdots \circ \phi_{2}\circ \phi_{1}$, where $K$ is the number of stages. After each stage $\phi_{k}$, $k=1,2,\cdots,K-1$, we insert an auxiliary branch $b_{k}$ for feature augmentation. Each $b_{k}$ includes a feature alignment module $\zeta_{k}$ and a linear classifier $g_{k}$. Given an input sample $\bm{x}$, the outputs of $K-1$ auxiliary branches and the backbone network $f$ are expressed as:

\begin{align}
&b_{1}(\bm{x})=g_{1}\circ \zeta _{1}\circ \phi_{1}(\bm{x}), \notag \\
&b_{2}(\bm{x})=g_{2}\circ \zeta _{2}\circ \phi_{2}\circ \phi_{1}(\bm{x}), \notag \\
&\cdots \notag \\
&b_{K-1}(\bm{x})=g_{K-1}\circ \zeta_{K-1}\circ \phi_{K-1}\circ \cdots \circ \phi_{1}(\bm{x}), \notag\\
& f(\bm{x})=g\circ \phi(\bm{x}).
\end{align}

Here, $b_{1}(\bm{x}),\cdots,b_{K-1}(\bm{x})\in \mathbb{R}^{C}$ and $f(\bm{x})\in \mathbb{R}^{C}$ are predicted logit vectors, where $C$ is the number of classes.  The feature alignment module $\zeta_{k}$ is to transform the feature map output from the shallow stage to match the feature dimension with the last stage. This is implemented by making each auxiliary branch's path from input to output have the same number of down-samplings as the backbone network. We formulate $F_{k}\in \mathbb{R}^{H\times W\times C}$ as the feature map output from the $k$-th branch $b_{k}$ for all $k=1,\cdots,K-1$ and $F_{K}\in \mathbb{R}^{H\times W\times C}$ as the feature map output from the backbone network $f$.  Here, $H,W$ and $C$ denote the height, width and channel number of the feature map, respectively. $\{F_{k}\}_{k=1}^{K-1}$ and $F_{K}$ are formulated as:
\begin{align}
F_{k}(\bm{x})=\zeta_{k}\circ \phi_{k}\circ \cdots \circ \phi_{1}(\bm{x}),\  F_{K}(\bm{x})=\phi(\bm{x}),
\end{align}
where $k=1,\cdots,K-1$. The transformed feature map often encodes global structures and coarse semantic information. The property is crucial for recognition performance by reducing redundant details in shallow features. \emph{The detailed architectures of auxiliary branches for various networks are shown in Appendix.}

\subsection{Task-Guided Classification Loss}
We describe how to construct classification loss from \emph{architecture source} and \emph{data source}. We illustrate the detailed overview of MixSKD in Fig.~\ref{MixSKD}.

\textbf{Classification loss from architecture source.} Given an input sample $\bm{x}$ with the ground-truth label $y$, we guide the $K-1$ auxiliary branches and backbone network $f$ to learn cross-entropy based classification loss:
\begin{equation}
L_{cls\_b\_f}(\bm{x},y)=L_{ce}(\sigma(f(\bm{x})),y)+\sum_{k=1}^{K-1}L_{ce}(\sigma(b_{k}(\bm{x})),y),
\end{equation}
where $\sigma$ represents a \emph{softmax} function to normalize the logits to model posterior probability distributions, and $L_{ce}$ denotes the cross-entropy. This conventional loss enables auxiliary branches and the backbone network to learn general classification capability and semantic features.

\textbf{Classification loss from data source.} Given two different images $\bm{x}_{i}$ and $\bm{x}_{j}$ with ground-truth labels $y_{i}$ and $y_{j}$, Mixup~\cite{zhang2017mixup} performs linear mixture
with a combination factor $\lambda$ to construct a virtual image $\tilde{\bm{x}}_{ij}$ with an interpolated one-hot ground-truth label $\tilde{y}_{ij}$:
\begin{align}
\tilde{\bm{x}}_{ij}=\lambda\bm{x}_{i}+(1-\lambda)\bm{x}_{j},\ \tilde{y}_{ij}=\lambda y_{i}+(1-\lambda)y_{j}.
\end{align}
Here, $\lambda\sim {\rm{Beta}}(\alpha,\alpha)$ for $\alpha\in (0,\infty)$ and $\lambda\in [0,1]$. We utilize the input images $\bm{x}_{i}$, $\bm{x}_{j}$ and $\tilde{\bm{x}}_{ij}$ to compute the classification task loss using cross-entropy:
\begin{align}
L_{cls\_mixup}=&L_{cls\_b\_f}(\bm{x}_{i},y_{i})+L_{cls\_b\_f}(\bm{x}_{j},y_{j})  +L_{cls\_b\_f}(\tilde{\bm{x}}_{ij},\tilde{y}_{ij})
\end{align}

\subsection{Feature map Self-KD}
Feature maps often contain information about image intensity and spatial correlation. The hidden feature maps often encode the intermediate learning process. We expect to encourage the network to behave consistently in \emph{intermediate feature maps} between the pair of original images $(\bm{x}_{i},\bm{x}_{j})$ and their \emph{Mixup} image $\tilde{\bm{x}}_{ij}$.  Thus we consider linearly interpolating the feature maps between $\bm{x}_{i}$ and $\bm{x}_{j}$. The interpolated feature maps are formulated as $\{\tilde{F}_{k}(\bm{x}_{i},\bm{x}_{j})\}_{k=1}^{K}$:
\begin{align}
\tilde{F}_{k}(\bm{x}_{i},\bm{x}_{j})=\lambda F_{k}(\bm{x}_{i})+(1-\lambda)F_{k}(\bm{x}_{j}).
\end{align}

Motivated by the hint loss in FitNet~\cite{romero2014fitnets}, we employ the squared $l_{2}$-norm for mutual alignment between interpolated feature maps $\{\tilde{F}_{k}(\bm{x}_{i},\bm{x}_{j})\in \mathbb{R}^{H\times W\times C}\}_{k=1}^{K}$ and \emph{Mixup} feature maps $\{F_{k}(\tilde{\bm{x}}_{ij})\in \mathbb{R}^{H\times W\times C}\}_{k=1}^{K}$:
\begin{equation}
L_{feature}=\sum_{k=1}^{K}\frac{1}{HWC}\parallel \tilde{F}_{k}(\bm{x}_{i},\bm{x}_{j}) - F_{k}(\tilde{\bm{x}}_{ij})\parallel^{2}. 
\label{b_hint}
\end{equation}
Inspired by Adversarial Feature Distillation (AFD)~\cite{wang2018adversarial}, we further introduce $K$ discriminators $\{D_{k}\}_{k=1}^{K}$ for the same-staged $K$ feature maps. The discriminator in the original AFD~\cite{wang2018adversarial} is used to distinguish features extracted from student or teacher. Instead, our discriminator is to classify the feature map generated from a linear interpolation or a Mixup image. The discriminator loss is formulated as a binary cross-entropy:
\begin{equation}
L_{dis}=-\sum_{k=1}^{K}\log[D_{k}(\tilde{F}_{k}(\bm{x}_{i},\bm{x}_{j}))]+\log[1-D_{k}(F_{k}(\tilde{\bm{x}}_{ij}))]. 
\label{b_dis}
\end{equation}
Here, the $D_{k}$ is a two-layer MLP followed by a Sigmoid function. Motivated by the idea of GAN~\cite{goodfellow2014generative}, jointly optimizing Eq.~(\ref{b_hint}) and Eq.~(\ref{b_dis}) via an adversarial process enables the network to learn common semantic information meaningfully. The feature Self-KD conducts an adversarial process to match linearly interpolated feature maps from the pair of original images $(\bm{x}_{i},\bm{x}_{j})$ with feature maps from the Mixup image $\tilde{\bm{x}}_{ij}$. This forces the network to behave linearly in feature representations between training samples.

\subsection{Logit Distribution Self-KD}
Beyond feature maps, logit-based class probability distributions often represent the final predictive information. We also expect to encourage the network to behave consistently in \emph{logit-based class posterior distributions} between the pair of original images $(\bm{x}_{i},\bm{x}_{j})$ and their Mixup image $\tilde{\bm{x}}_{ij}$. The linear interpolations of images should lead to linear interpolations of corresponding logit-based probability distributions with the same mixed proportion. The interpolated logits vectors of $K-1$ auxiliary branches and the backbone network $f$ are formulated as $\{\tilde{b}_{k}(\bm{x}_{i},\bm{x}_{j})\}_{k=1}^{K-1}$ and $\tilde{f}(\bm{x}_{i},\bm{x}_{j})$:

\begin{equation}
\tilde{b}_{k}(\bm{x}_{i},\bm{x}_{j})=\lambda b_{k}(\bm{x}_{i})+(1-\lambda)b_{k}(\bm{x}_{j}),\ \tilde{f}(\bm{x}_{i},\bm{x}_{j})=\lambda f(\bm{x}_{i})+(1-\lambda)f(\bm{x}_{j}).
\end{equation}
\subsubsection{Logit Self-KD over auxiliary branches $\{b_{k}\}_{k=1}^{K-1}$.} 
First, we aim to maximize the consistency of logit-based class probability distributions between linearly interpolated logits $\{\tilde{b}_{k}(\bm{x}_{i},\bm{x}_{j})\}_{k=1}^{K-1}$ and Mixup logits $\{b_{k}(\tilde{\bm{x}}_{ij})\}_{k=1}^{K-1}$ via KL-divergence: 
\begin{align}
L_{b\_logit}((\bm{x}_{i},\bm{x}_{j}),\tilde{\bm{x}}_{ij})
=\sum_{k=1}^{K-1}[&L_{KL}(\sigma(\tilde{b}_{k}(\bm{x}_{i},\bm{x}_{j})/T),\sigma(\overline{b_{k}(\tilde{\bm{x}}_{ij})}/T)) \notag \\
+&L_{KL}(\sigma(b_{k}(\tilde{\bm{x}}_{ij})/T),\sigma(\overline{\tilde{b}_{k}(\bm{x}_{i},\bm{x}_{j})}/T)]. 
\label{blogit}
\end{align}
Here, $L_{KL}$ is the KL-divergence, and $T$ is a temperature following the original KD~\cite{hinton2015distilling}. $\overline{b_{k}(\tilde{\bm{x}}_{ij})}$ and $\overline{\tilde{b}_{k}(\bm{x}_{i},\bm{x}_{j})}$ are fixed copies of $b_{k}(\tilde{\bm{x}}_{ij})$ and $\tilde{b}_{k}(\bm{x}_{i},\bm{x}_{j})$, respectively. As suggested by~\cite{miyato2018virtual}, the gradients through  $\overline{b_{k}(\tilde{\bm{x}}_{ij})}$ and $\overline{\tilde{b}_{k}(\bm{x}_{i},\bm{x}_{j})}$ are not propagated to avoid the model collapse issue. 

\subsubsection{Training the self-teacher network $h(\cdot)$.} The idea of the self-teacher $h(\cdot)$ is to aggregate multi-stage ensemble feature maps to construct an excellent classifier. The classifier could provide meaningful soft labels to teach the backbone classifier in an online manner. Given the $K$ feature maps $\{\mathcal{F}_{k}\}_{k=1}^{K}$ with the same dimension to $h(\cdot)$, we first concatenate them along the channel dimension and then use a $1\times 1$ convolution $\rm{Conv}_{1\times 1}$ to shrink  channels. Then the ensemble feature maps are followed by a linear classifier $g_{h}$ to output class probability distributions. The inference graph is formulated as $h(\{\mathcal{F}_{k}\}_{k=1}^{K})=g_{h}({\rm{Conv}_{1\times 1}}([\mathcal{F}_{1}, \cdots, \mathcal{F}_{K}]))$.

Given the linearly interpolated feature maps $\{\tilde{F}_{k}(\bm{x}_{i},\bm{x}_{j})\}_{k=1}^{K}$ from the pair of original images $(\bm{x}_{i},\bm{x}_{j})$, we can derive the soft label $h(\{\tilde{F}_{k}(\bm{x}_{i},\bm{x}_{j})\}_{k=1}^{K})$. Given Mixup feature maps  $\{F_{k}(\tilde{\bm{x}}_{ij})\}_{k=1}^{K}$ from the Mixup image $\tilde{\bm{x}}_{ij}$, we can derive the soft label $h(\{F_{k}(\tilde{\bm{x}}_{ij})\}_{k=1}^{K})$. For easy notation, we define $h(\bm{x}_{i},\bm{x}_{j})=h(\{\tilde{F}_{k}(\bm{x}_{i},\bm{x}_{j})\}_{k=1}^{K})$ and $h(\tilde{\bm{x}}_{ij})=h(\{F_{k}(\tilde{\bm{x}}_{ij})\}_{k=1}^{K})$. Motivated by the success of Manifold Mixup~\cite{verma2019manifold}, the class probabilities inferred from linear interpolated feature maps could also be supervised by interpolated labels. Thus we further train both $h(\bm{x}_{i},\bm{x}_{j})$ and $h(\tilde{\bm{x}}_{ij})$ by the interpolated label $\tilde{y}_{ij}$ using cross-entropy loss:
\begin{equation}
L_{cls\_h}=L_{ce}(\sigma(h(\bm{x}_{i},\bm{x}_{j})),\tilde{y}_{ij})+L_{ce}(\sigma(h(\tilde{\bm{x}}_{ij})),\tilde{y}_{ij}).
\label{cls_h}
\end{equation}
Benefiting from Equ.~(\ref{cls_h}), $h(\cdot)$ would learn meaningful mixed distributions from multi-stage feature maps from the view of information mixture.

\subsubsection{Logit Self-KD over the backbone $f$.} 
We adopt the self-teacher network $h(\cdot)$ to supervise the final backbone classifier, since $h(\cdot)$ aggregates multi-branch feature information and leads to better performance. We utilize the Mixup logit $h(\tilde{\bm{x}}_{ij})$ as the soft label to supervise the linearly interpolated logit $\tilde{f}(\bm{x}_{i},\bm{x}_{j})$. We also employ logit $h(\bm{x}_{i},\bm{x}_{j})$ from the linearly interpolated feature map to supervise Mixup logit $f(\tilde{\bm{x}}_{ij})$. The mutual distillation loss is formulated as:
\begin{align}
L_{f\_logit}((\bm{x}_{i},\bm{x}_{j}),\tilde{\bm{x}}_{ij}) 
&=L_{KL}(\sigma(\tilde{f}(\bm{x}_{i},\bm{x}_{j})/T),\sigma(h(\tilde{\bm{x}}_{ij})/T))\notag \\
&+L_{KL}(\sigma(f(\tilde{\bm{x}}_{ij})/T),\sigma(h(\bm{x}_{i},\bm{x}_{j})/T)).  
\label{flogit}
\end{align}

Here, the gradients through the soft labels of $h(\tilde{\bm{x}}_{ij})$ and $h(\bm{x}_{i},\bm{x}_{j})$ are not propagated. 
In theory, the loss  $L_{b\_logit}$ and $L_{f\_logit}$ can regularize the consistency between linearly interpolated class probability distributions from the pair of original images $(\bm{x}_{i},\bm{x}_{j})$ and the distribution from Mixup image $\tilde{\bm{x}}_{ij}$. This encourages the network to make linear predictions in-between training samples over auxiliary branches $\{b_{k}\}_{k=1}^{K-1}$ and backbone network $f$.

\subsection{Overall Loss of MixSKD}
We summarize the loss terms above into a unified framework:
\begin{align}
L_{MixSKD}=&\underbrace{L_{cls\_mixup}}_{task\ loss}+\underbrace{\beta L_{feature}+\gamma L_{dis}}_{feature\ Self-KD} 
+\mu(\underbrace{L_{b\_logit}+L_{cls\_h}+L_{f\_logit}}_{logit\ Self-KD}).
\label{overall}
\end{align}
Here, we use $\beta=1$ to control the magnitude of feature $l_{2}$ loss, where we find $\beta\in [1, 10]$ works well. Besides, we choose $\gamma=1$ for discriminator loss and $\mu=1$ for cross-entropy or KL divergence losses since they are probability-based forms and in the same magnitude. We perform end-to-end optimization of the training graph. During the test stage, we discard attached auxiliary components, leading to no extra inference costs compared with the original network.

In theory, our proposed MixSKD encourages the network to behave linearly in the latent feature and probability distribution spaces. The linear behaviour may reduce undesirable oscillations when predicting outliers, as discussed by Zhang \emph{et al.}~\cite{zhang2017mixup}. Moreover, linearity is also an excellent inductive bias from the view of Occam's razor because it is one of the most straightforward behaviours.
\section{Experiments}
\subsection{Experimental Setup}
\textbf{Dataset.} We conduct experiments on CIFAR-100~\cite{krizhevsky2009learning} and ImageNet~\cite{deng2009imagenet} as the standard image classification tasks and CUB-200-2011 (CUB200)~\cite{wah2011caltech}, Standford
Dogs (Dogs)~\cite{khosla2011novel}, MIT Indoor Scene Recognition (MIT67)~\cite{quattoni2009recognizing}, Stanford Cars (Cars)~\cite{krause20133d} and FGVC-Aircraft (Air)~\cite{maji2013fine} datasets as the fine-grained classification tasks. For downstream dense prediction tasks, we use COCO 2017~\cite{lin2014microsoft} for object detection and Pascal VOC~\cite{everingham2010pascal}, ADE20K~\cite{zhou2017scene} and COCO-Stuff-164K~\cite{caesar2018coco} for semantic segmentation. \emph{The detailed dataset descriptions and training details are provided in the Appendix.} We report the mean accuracy with a standard deviation over three runs using the form of $mean_{\pm std}$ for CIFAR-100 and fine-grained classification. 

\textbf{Hyper-parameter setup.} As suggested Mixup~\cite{zhang2017mixup}, we use $\alpha=0.2$ in MixSKD for ImageNet and $\alpha=0.4$ for other image classification datasets. Moreover, we set temperature $T=3$ for all datasets.

\subsection{Comparison with State-of-the-arts}
\begin{table}[tbp]
	\centering
	\caption{Top-1 accuracy(\%) of various Self-KD (the second block) and  data augmentation (the third block) methods across widely used networks for CIFAR-100 classification. The numbers in \textbf{bold} and \underline{underline} indicate \textbf{the best} and \underline{the second-best} results, respectively. * denotes the result from the original paper.}
	\begin{tabular}{l|cccc}  
		\toprule
		Method&ResNet-18\cite{he2016deep}&WRN-16-2\cite{zagoruyko2016wide}&DenseNet-40\cite{huang2017densely}&HCGNet-A1\cite{yang2020gated} \\ 
		Parameters&11.2M&0.7M&1.1M&1.1M \\
		\midrule
		Baseline & 76.24$_{\pm 0.07}$ & 72.24$_{\pm 0.29}$ &74.61$_{\pm 0.24}$ & 75.58$_{\pm 0.30}$\\
		\midrule
		DDGSD~\cite{xu2019data}&76.61$_{\pm 0.47}$&72.46$_{\pm 0.05}$ & 75.87$_{\pm 0.30}$&76.50$_{\pm 0.18}$\\
		DKS~\cite{sun2019deeply}& 78.64$_{\pm 0.25}$ &73.73$_{\pm 0.22}$ &74.94$_{\pm 0.60}$&78.04$_{\pm 0.16}$ \\
		BYOT~\cite{zhang2019your}& 77.88$_{\pm 0.19}$ & 72.97$_{\pm 0.34}$&75.49$_{\pm 0.16}$&76.81$_{\pm 0.64}$ \\
		SAD~\cite{hou2019learning}&76.40$_{\pm 0.17}$ &72.62$_{\pm 0.17}$ &74.77$_{\pm 0.12}$&75.86$_{\pm 0.57}$ \\
		Tf-KD~\cite{yuan2020revisiting}& 76.61$_{\pm 0.34}$& 72.66$_{\pm 0.21}$&74.68$_{\pm 0.17}$&75.84$_{\pm 0.23}$ \\
		CS-KD~\cite{yun2020regularizing}& 78.01$^{*}$$_{\pm 0.13}$ &73.23$_{\pm 0.33}$ & 75.02$_{\pm 0.37}$&76.36$_{\pm 0.30}$\\
		FRSKD~\cite{ji2021refine}& 77.71$^{*}$$_{\pm 0.14}$&73.27$^{*}$$_{\pm 0.45}$ & 74.91$_{\pm 0.36}$&77.26$_{\pm 0.11}$\\
		\midrule
		Cutout~\cite{devries2017improved} & 76.66$_{\pm 0.42}$ & 73.66$_{\pm 0.18}$ &75.45$_{\pm 0.33}$&76.63$_{\pm 0.15}$ \\
		Mixup~\cite{zhang2017mixup} & 78.68$_{\pm 0.12}$ & 73.60$_{\pm 0.59}$ &75.55$_{\pm 0.81}$&77.89$_{\pm 0.18}$ \\
		Manifold Mixup~\cite{verma2019manifold} & 79.29$_{\pm 0.20}$ & 72.53$_{\pm 0.08}$ &75.19$_{\pm 0.38}$&77.72$_{\pm 0.26}$ \\
		AutoAugment~\cite{cubuk2019autoaugment} & 77.97$_{\pm 0.17}$ & \underline{74.16}$_{\pm 0.22}$ &76.21$_{\pm 0.15}$ &78.13$_{\pm 0.78}$\\
		RandAugment~\cite{cubuk2020randaugment} & 76.86$_{\pm 0.55}$ & 73.87$_{\pm 0.08}$ &76.23$_{\pm 0.22}$&77.56$_{\pm 0.50}$ \\
		Random Erase~\cite{zhong2020random} & 76.75$_{\pm 0.33}$ & 74.04$_{\pm 0.11}$ &75.52$_{\pm 0.46}$&77.38$_{\pm 0.14}$ \\
		\midrule
		CS-KD+Mixup~\cite{yun2020regularizing} & \underline{79.60$^{*}$}$_{\pm 0.31}$ & 73.82$_{\pm 0.18}$ &\underline{76.32}$_{\pm 0.37}$ &78.03$_{\pm 0.44}$\\
		FRSKD+Mixup~\cite{ji2021refine} & 78.74$^{*}$$_{\pm 0.19}$ & 73.67$_{\pm 0.36}$ &75.56$_{\pm 0.07}$&\underline{78.23}$_{\pm 0.22}$ \\
		\midrule
		MixSKD (Ours)& \textbf{80.32}$_{\pm 0.13}$ & \textbf{74.89}$_{\pm 0.27}$ &\textbf{76.85}$_{\pm 0.19}$&\textbf{78.57}$_{\pm 0.19}$ \\
		\bottomrule
	\end{tabular}
	
	\label{cifar100} 
\end{table}

\textbf{Performance comparison on CIFAR-100.} Table~\ref{cifar100} shows the accuracy comparison towards state-of-the-art data augmentation and Self-KD methods on CIFAR-100. Because MixSKD integrates Mixup~\cite{zhang2017mixup}, we also compare our method with the latest Self-KD methods of CS-KD~\cite{yun2020regularizing} and FRSKD~\cite{ji2021refine} combined with Mixup. We use ResNet~\cite{he2016deep}, WRN~\cite{zagoruyko2016wide}, DenseNet~\cite{huang2017densely} and HCGNet~\cite{yang2020gated} as backbone networks to evaluate the performance. Almost previous data augmentation and Self-KD methods can enhance the classification performance upon the baseline across various networks. Compared to the Mixup, MixSKD achieves an average accuracy gain of 1.23\% across four networks. The result indicates that the superiority of MixSKD is not only attributed to the usage of Mixup. We further demonstrate that MixSKD is superior to the state-of-the-art CS-KD+Mixup and FRSKD+Mixup with average margins of 0.71\% and 1.11\%, respectively. These results imply that our MixSKD explores more meaningful knowledge from image mixture than the conventional training signals for Self-KD. Moreover, we find that it is hard to say which is
the second-best approach since different methods are superior
for various architectures. MixSKD further outperforms the powerful data augmentation method AutoAugment with an average gain of 1.04\%.

\textbf{Performance comparison on fine-grained classification.} Compared with standard classification, fine-grained classification often contains
fewer training samples per class and more similar inter-class semantics. This challenges a network to learn more discriminative intra-class variations. As
shown in Table~\ref{finegrained}, we train a ResNet-18 on five fine-grained classification tasks. MixSKD can also achieve the best performance on fine-grained classification. MixSKD outperforms the best-second results with 2.86\%, 0.50\%, 1.99\%, 2.07\% and 1.82\% accuracy gains on five datasets from left to right. The results verify that MixSKD can regularize the network to capture more discriminative features.

\begin{table}[tbp]
	\centering
	\caption{Top-1 accuracy(\%) of various Self-KD methods on ResNet-18 for fine-grained classification. The numbers in \textbf{bold} and \underline{underline} indicate \textbf{the best} and \underline{the second-best} results, respectively. * denotes the result from the original paper. }
	\begin{tabular}{l|ccccc}  
		\toprule
		Method&CUB200&Dogs& MIT67& Cars&Air \\ 
		\midrule
		Baseline & 57.48$_{\pm 0.45}$ & 66.83$_{\pm 0.29}$ &57.81$_{\pm 1.42}$&83.50$_{\pm 0.24}$&77.07$_{\pm 0.26}$ \\
		DDGSD~\cite{xu2019data}&56.89$_{\pm 0.42}$&69.24$_{\pm 0.84}$ & 56.46$_{\pm 0.59}$&85.04$_{\pm 0.11}$&74.91$_{\pm 0.97}$\\
		DKS ~\cite{sun2019deeply}& 63.72$_{\pm 0.21}$ &71.07$_{\pm 0.07}$ &61.50$_{\pm 0.12}$&86.13$_{\pm 0.31}$&79.69$_{\pm 0.31}$ \\
		BYOT~\cite{zhang2019your}& 61.77$_{\pm 0.43}$ & 69.58$_{\pm 0.20}$&59.03$_{\pm 0.42}$&85.36$_{\pm 0.18}$&79.32$_{\pm 0.45}$ \\
		SAD~\cite{hou2019learning}&55.51$_{\pm 0.67}$ &66.10$_{\pm 0.08}$ &57.46$_{\pm 0.79}$&82.94$_{\pm 0.22}$&73.62$_{\pm 0.68}$ \\
		Tf-KD~\cite{yuan2020revisiting}& 57.44$_{\pm 0.25}$& 66.57$_{\pm 0.33}$&57.51$_{\pm 0.86}$&83.59$_{\pm 0.49}$&76.76$_{\pm 0.34}$ \\
		CS-KD~\cite{yun2020regularizing}& 66.72$^{*}$$_{\pm 0.99}$ &69.15$^{*}$$_{\pm 0.28}$ & 59.55$^{*}$$_{\pm 0.45}$&86.87$_{\pm 0.04}$&80.92$_{\pm 0.44}$\\
		FRSKD~\cite{ji2021refine}& 65.39$^{*}$$_{\pm 0.13}$&70.77$^{*}$$_{\pm 0.20}$ & 61.74$^{*}$$_{\pm 0.67}$&84.73$_{\pm 0.03}$&78.85$_{\pm 0.55}$\\
		\midrule
		Mixup~\cite{zhang2017mixup} & 65.53$_{\pm 0.73}$ & 69.30$_{\pm 0.10}$ &58.83$_{\pm 0.77}$&86.10$_{\pm 0.28}$&79.94$_{\pm 0.18}$ \\
		CS-KD+Mixup & \underline{69.29}$^{*}$$_{\pm 0.64}$ & 70.07$^{*}$$_{\pm 0.14}$ &60.35$^{*}$$_{\pm 0.85}$&\underline{87.10}$_{\pm 0.30}$&\underline{81.13}$_{\pm 0.45}$ \\
		FRSKD+Mixup & 67.98$^{*}$$_{\pm 0.58}$ & \underline{71.64}$^{*}$$_{\pm 0.29}$ &\underline{62.11}$^{*}$$_{\pm 0.81}$&86.25$_{\pm 0.33}$&79.97$_{\pm 0.58}$ \\
		\midrule
		MixSKD (Ours)& \textbf{72.15}$_{\pm 0.53}$ & \textbf{72.14}$_{\pm 0.22}$ &\textbf{64.10}$_{\pm 0.45}$&\textbf{89.17}$_{\pm 0.08}$&\textbf{82.95}$_{\pm 0.31}$ \\
		\bottomrule
	\end{tabular}
	
	\label{finegrained} 
\end{table}

\textbf{Performance comparison on ImageNet classification.} ImageNet is a large-scale image classification dataset, which is a golden classification benchmark. As shown in Table~\ref{imagenet}, MixSKD achieves the best $78.76\%$ and $94.40\%$ top-1 and top-5 accuracies on ResNet-50 compared to advanced data augmentation and Self-KD methods. It surpasses the best-competing FRSKD+Mixup by $0.97\%$ and $0.76\%$ top-1 and top-5 accuracy gains. The results show the scalability of our MixSKD to work reasonably well on the large-scale dataset.

\textbf{Performance comparison on transfer learning to object detection and semantic segmentation.} We use the ResNet-50 pre-trained on ImageNet as a backbone over
Cascade R-CNN~\cite{cai2019cascade} to perform object detection on COCO 2017~\cite{lin2014microsoft} and over DeepLabV3~\cite{chen2017rethinking} to perform semantic segmentation on Pascal VOC~\cite{everingham2010pascal}, ADE20K~\cite{zhou2017scene} and COCO-Stuff-164K~\cite{caesar2018coco}. We follow the standard data preprocessing~\cite{ren2016faster,yang2022cross} and implement object detection over MMDetection~\cite{chen2019mmdetection} with 1x training schedule and semantic segmentation over an open codebase released by Yang~\emph{et al.}~\cite{yang2022cross}. As shown in Table~\ref{imagenet}, MixSKD achieves the best downstream performance and outperforms the best-second results with $0.4\%$ mAP on detection and $1.68\%$, $2.65\%$ and $3.04\%$ mIoU margins on Pascal VOC, ADE20K and COCO-Stuff segmentation, respectively. In contrast, although other state-of-the-art Self-KD and data augmentation methods achieve good performance on ImageNet, they often transfer worse to downstream recognition tasks. The results
indicate that our method can guide the network to learn better transferable feature representations for downstream dense prediction tasks.

\begin{table}[t]
	\centering
	\caption{Performance comparison with data augmentation and Self-KD methods on ImageNet and downstream transfer learning for object detection on COCO 2017 and semantic segmentation on Pascal VOC, ADE20K and COCO-Stuff-164K. All results are re-implemented by ourselves.  The numbers in \textbf{bold} and \underline{underline} indicate \textbf{the best} and \underline{the second-best} results, respectively.}
	\resizebox{1.\linewidth}{!}{
		\begin{tabular}{l|cc|c|ccc}  
			\toprule 
			
			\multirow{3}{*}{Method}&\multicolumn{2}{c|}{Image Classification} &Detection&\multicolumn{3}{c}{Semantic Segmentation} \\ \cline{2-7}
			&\multicolumn{2}{c|}{ImageNet} &COCO&Pascal VOC& ADE20K&COCO-Stuff \\ 
			&Top-1 Acc(\%)&Top-5 Acc(\%)&mAP(\%)&\multicolumn{3}{c}{mIoU(\%)} \\  
			\midrule
			ResNet-50&77.08&93.20& 41.0&77.09&\underline{39.72}&\underline{34.08} \\
			+Mixup~\cite{zhang2017mixup}&77.51&\underline{93.72}& \underline{41.1}&76.71&39.69&33.38 \\
			+Manifold Mixup~\cite{verma2019manifold}&77.43&93.68& 40.8&75.62&38.07&32.82 \\
			+FRSKD~\cite{ji2021refine}&76.92&92.96& 40.9&\underline{77.10}&39.48&33.63 \\
			+FRSKD+Mixup~\cite{ji2021refine}&\underline{77.79}&93.64& 41.0&76.61&39.39&33.54 \\
			+MixSKD (Ours)&\textbf{78.76}&\textbf{94.40} &\textbf{41.5}&\textbf{78.78}&\textbf{42.37}&\textbf{37.12} \\
			\bottomrule
	\end{tabular}}
	\label{imagenet}
\end{table}

\begin{table}[t]
	\centering
	\caption{Ablation study of loss terms in MixSKD. The first column with all '-' denotes the baseline. All results denote top-1 accuracy over ResNet-50 on ImageNet.}
	\begin{tabular}{c|l|cccccccc}  
		\toprule
		Type&Loss&\multicolumn{8}{c}{Various loss combinations of MixSKD} \\ 
		\midrule
		Task&$L_{cls\_mixup}$&-&\checkmark&\checkmark&\checkmark&\checkmark&\checkmark&\checkmark&\checkmark \\
		\hline
		\multirow{2}{*}{Feature Self-KD}&$L_{feature}$&-&-&\checkmark &\checkmark&-&-&-&\checkmark\\
		&$L_{dis}$ &-&-&-&\checkmark&-&-&-&\checkmark\\
		\hline
		\multirow{3}{*}{Logit Self-KD}&$L_{b\_logit}$&-&-&-&-&\checkmark&-&\checkmark&\checkmark \\
		&$L_{cls\_h}$&-&-&-&-&-&\checkmark&\checkmark&\checkmark \\
		&$L_{f\_logit}$&-&-&-&-&-&\checkmark&\checkmark&\checkmark \\
		\midrule
		\multicolumn{2}{c|}{Top-1 Accuracy (\%)} &77.08&77.68&78.14&78.47&78.06&78.23&78.54&\textbf{78.76}\\
		\bottomrule
	\end{tabular}
	
	\label{attack}
\end{table}

\begin{table}[t]
	\caption{\textbf{\emph{Left}}: top-1 accuracy (\%) under FGSM white-box attack with various perturbation weights $\epsilon$ on CIFAR-100.  \textbf{\emph{Middle}}: top-1 accuracy (\%) under sensitivity analysis of temperature $T$ over ResNet-50 on ImageNet. \textbf{\emph{Right}}:  top-1 accuracy (\%) under ablation study of using different feature maps from various stages to construct the self-teacher $h$ over ResNet-50 on ImageNet.}
	
	\begin{minipage}{0.66\linewidth}
		\centering
		
		\resizebox{1.\linewidth}{!}{
			\begin{tabular}{l|cccc}  
				\toprule 
				
				Method&$\epsilon=0.0001$&$\epsilon=0.001$&$\epsilon=0.01$&$\epsilon=0.1$\\  
				\midrule
				WRN-16-2&60.5&57.9&37.6&8.6 \\
				+Mixup&63.1&61.1&40.7&11.4\\
				+CS-KD+Mixup&63.8&62.1&42.8&13.0 \\
				+FRSKD+Mixup&62.9&60.7&40.9&12.2 \\
				+MixSKD (Ours)&\textbf{67.7}&\textbf{65.5}&\textbf{46.8}&\textbf{15.3} \\
				\bottomrule
		\end{tabular}}
	\end{minipage}
	\hfill
	\begin{minipage}{0.12\linewidth}
		\centering
		\resizebox{1\textwidth}{!}{
			\begin{tabular}{cc}
				\toprule
				$T$ & Acc \\ \midrule 
				1 & 77.94 \\
				2 & 78.21 \\
				3 & \textbf{78.76} \\
				4 & 78.65 \\
				5 & 78.37\\\bottomrule
			\end{tabular}
		}
	\end{minipage}
	\hfill
	\begin{minipage}{0.165\linewidth}  
		\centering
		\resizebox{1\textwidth}{!}{
			\begin{tabular}{cc}
				\toprule
				Stage & Acc \\ \midrule 
				1& 78.01\\
				2 & 78.18  \\
				3 & 78.36 \\
				4 & 78.42 \\
				All & \textbf{78.76} \\ \bottomrule
			\end{tabular}
		}
	\end{minipage}
	
	\label{ablation}
\end{table}

\subsection{Ablation Study and Analysis}
\textbf{Ablation study of loss terms.} As shown in Table~\ref{attack}, we conduct thorough ablation experiments to evaluate the effectiveness of each component in MixSKD on the convincing ImageNet dataset:

 \textbf{(1) Task loss.} We regard the $\emph{L}_{cls\_mixup}$ as the basic task loss that trains the network with auxiliary architectures using Mixup. It improves the baseline by a 0.60\% accuracy gain. Moreover, $\emph{L}_{cls\_mixup}$ can guide the network to produce meaningful feature maps and logits from the original images and Mixup images. The information further motivates us to perform Self-KD to regularize the network. \textbf{(2) Feature map Self-KD.} Using $L_{feature}$ loss to distill $K$ feature maps achieves a $0.46\%$ improvement over the task loss. Adding an auxiliary discriminative loss $L_{dis}$ further enhances the performance of $L_{feature}$ with an extra $0.33\%$ gain. The result suggests that using an adversarial mechanism for feature distillation may encourage the network to capture more meaningful semantics than a single $l_{2}$-based mimicry loss. \textbf{(3) Logit Self-KD.} The distillation loss $L_{b\_logit}$ over auxiliary branches $\{b_{k}\}_{k=1}^{K-1}$ results in a $0.38\%$ gain than the task loss. Training the self-teacher network $h$ by $L_{cls\_h}$ for providing soft labels to supervise the backbone classifier $f$ by $L_{f\_logit}$ leads to a significant $0.55\%$ gain than the task loss. Overall, combining logit-based losses can enable the network to achieve a $0.86\%$ gain. \textbf{(4) Overall loss.} Combining logit and feature Self-KD can maximize the performance gain with a $1.08\%$ margin over the task loss.

\textbf{Adversarial Robustness.} Adversarial learning has shown that adding imperceptibly small but intentionally worst-case perturbations to the
input image fools the network easily. Thus adversarial robustness is a crucial issue for the practical application of neural networks. As shown in Table~\ref{ablation} (left), we compare the robustness trained by various Self-KD methods under FGSM white-box attack~\cite{goodfellow2014explaining}. MixSKD significantly improves the robustness and surpasses other methods consistently across different perturbations. The results imply that MixSKD shows better potential in adversarial defense.

\textbf{Impact of the temperature $T$.} The temperature $T$ is utilized to scale the predictive logits for KL-divergence based distillation losses in Eq.~(\ref{blogit}) and Eq.~(\ref{flogit}). A larger $T$ leads to a smoother probability distribution. As shown in Table~\ref{ablation}  (middle), we observe that MixSKD benefits most from $T=3$ than others for producing soft distributions.

\textbf{Effectiveness of multi-stage feature aggregation to construct self-teacher $h$.} The self-teacher utilizes feature maps not only from the last stage but also from multiple stages to aggregate richer semantics. As shown in Table~\ref{ablation} (right), we demonstrate that multi-stage feature aggregation can generate higher-quality soft labels to achieve better performance than any single feature map.  

\begin{figure}[tbp] 
	\centering 
	\subfigure[Comparison of log probabilities
	of predicted labels (\emph{left}) and ground-truth labels (\emph{right}) over common misclassified samples trained by baseline and MixSKD.]{  
		\label{misclassified_sample} 
		\includegraphics[width=0.65\textwidth]{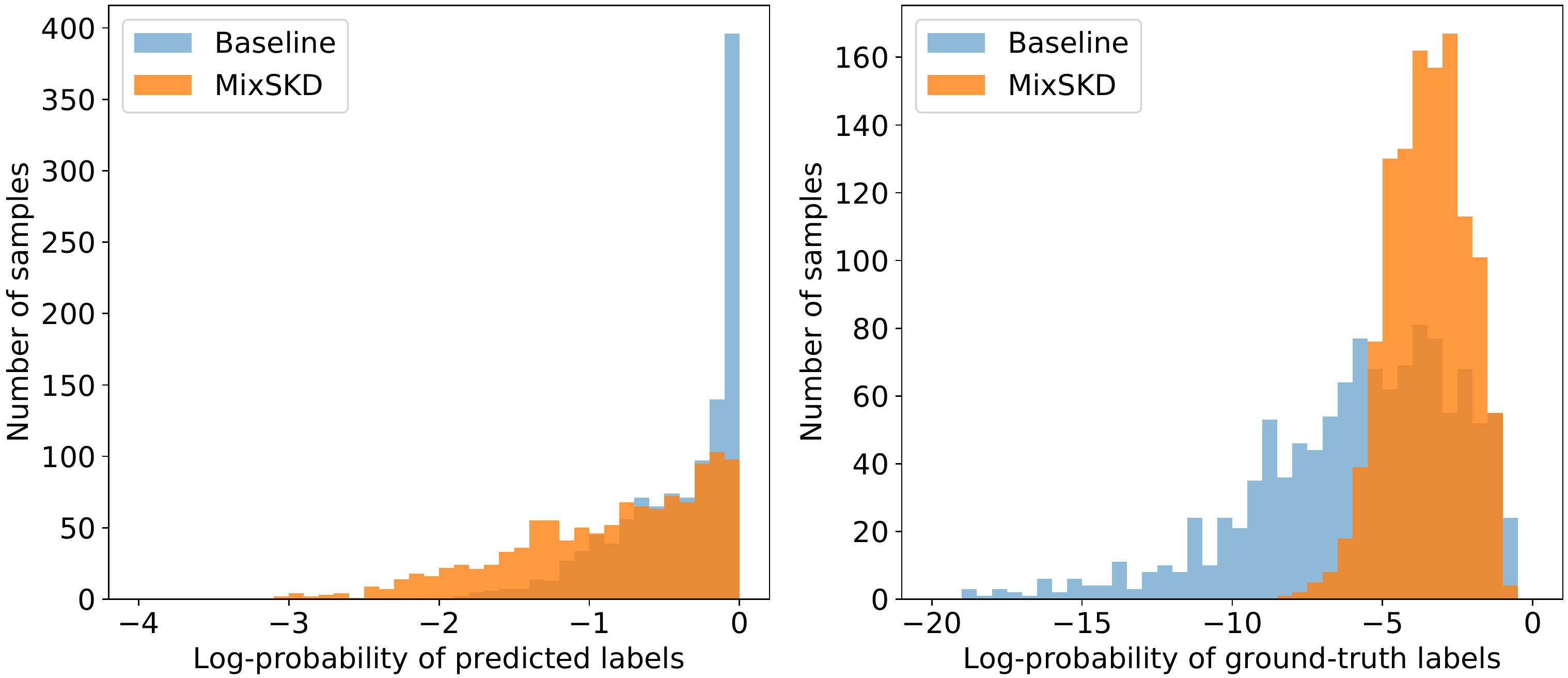} }
	\subfigure[Miss rate of the prediction of mixup image $\tilde{\bm{x}}_{ij}$ not belonging to $\{y_{i},y_{j}\}$. ]{ 
		\label{miss} 
		\includegraphics[width=0.315\textwidth]{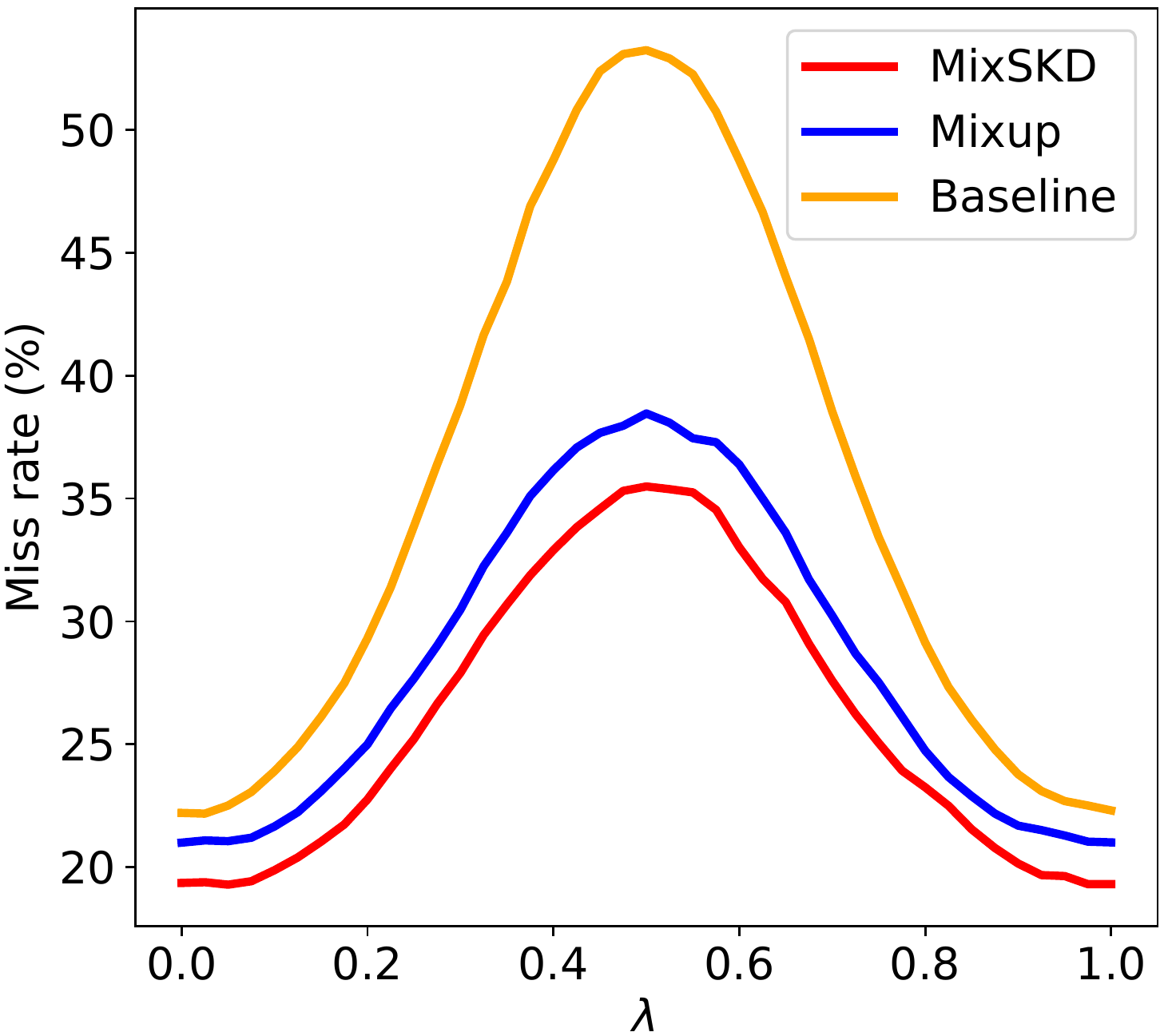}} 
	\caption{MixSKD produces more robust predictions on CIFAR-100.} 
	\label{Fig.lable} 
\end{figure}

\textbf{Meaningful predictive distribution.} By encouraging linear behaviours between samples, we examine whether our MixSKD forces the network to achieve a meaningful predictive distribution. To this end, we investigate predictive softmax scores inferred from ResNet-18 trained by baseline and MixSKD  on CIFAR-100. Fig.~\ref{misclassified_sample} shows the histogram statistics of log probabilities of predicted labels and ground-truth labels over common misclassified samples. Compared to the baseline, our MixSKD achieves better prediction quality. On the one hand, it decreases over-confident log probabilities of predicted labels over misclassified samples effectively. On the other hand, it assigns larger log probabilities for ground-truth labels even if the misclassification occurs.

\textbf{Miss error rate in-between samples.} As shown in Fig.~\ref{miss}, we evaluate the network trained by baseline, Mixup, and our MixSKD on the Mixup image $\tilde{\bm{x}}_{ij}=\lambda \bm{x}_{i}+(1-\lambda) \bm{x}_{j}$. A prediction is regarded as "miss" if it does not belong to $\{y_{i},y_{j}\}$. Our MixSKD shows the minimum miss rates across various Mixup factors of $\lambda$ within the range of $[0,1]$. The results indicate that MixSKD leads to more robust predictive behaviours in-between samples.

\section{Conclusion}
This paper presents MixSKD, a powerful Self-KD method to regularize the network to behave linearly in feature maps and class probabilities between samples using Mixup images. Moreover, we also construct an auxiliary self-teacher to transfer external ensemble knowledge for the backbone network. The overall MixSKD outperforms state-of-the-art data augmentation and Self-KD methods on computer vision benchmarks. We hope this
work will inspire further research towards information mixture to improve the performance of visual recognition.

\noindent\textbf{Acknowledgment.} Zhulin An is the corresponding author.
\clearpage
%
%
\bibliographystyle{splncs04}
\bibliography{egbib}

\appendix 

\section{Architectural Design of Auxiliary Branches}
As discussed in the main paper, we attach one auxiliary branch $b_{k}$ after each convolutional stage $\phi_{k},k=1,2,\cdots,K-1$. Each auxiliary branch $b_{k}$ includes a feature alignment module $\zeta_{k}$ and a linear classifier $g_{k}$.
The feature alignment module contains several convolutional blocks following the original backbone network, \emph{i.e.}
residual block in ResNets~\cite{he2016deep}. To enable the fine-to-coarse feature transformation, we make the path from the input to the end of each auxiliary branch $b_{k}$ have the same number of down-sampling as the backbone network $f$. We illustrate the overall architectures of various networks with auxiliary branches involved in the main paper, including ResNet~\cite{he2016deep}, WRN~\cite{zagoruyko2016wide}, DenseNet~\cite{huang2017densely} and HCGNet~\cite{yang2020gated}. For better readability, the style of the illustration of
architectural details is followed by the original paper.

\begin{table}
	\caption{Architectural details of the backbone ResNet-18~\cite{he2016deep} with auxiliary branches for CIFAR-100 classification.}
	\centering
	\resizebox{1\linewidth}{!}{
		\begin{tabular}{c|c|c|c|c|c}
			\hline
			Layer name & Output size & $f(\cdot)$& $b_{1}(\cdot)$& $b_{2}(\cdot)$& $b_{3}(\cdot)$\\ 	\hline
			conv1&  32$\times$32 &  $3\times3, 64$ &-&-&-\\  \hline
			conv2\_x &  32$\times$32              & 
			$\begin{bmatrix}
			3\times 3, 64\\
			3\times 3, 64
			\end{bmatrix}\times 2 $  &  -&  - &  - \\  \hline
			conv3\_x &  16$\times$16           & 
			$\begin{bmatrix}
			3\times 3, 128\\
			3\times 3, 128
			\end{bmatrix}\times 2 $  &  $\begin{bmatrix}
			3\times 3, 128\\
			3\times 3, 128
			\end{bmatrix}\times 2 $ & - & -  \\  \hline
			
			conv4\_x &  8$\times$8           & 
			$\begin{bmatrix}
			3\times 3, 256\\
			3\times 3, 256
			\end{bmatrix}\times 2 $  &  $\begin{bmatrix}
			3\times 3, 256\\
			3\times 3, 256
			\end{bmatrix}\times 2 $&  $\begin{bmatrix}
			3\times 3, 256\\
			3\times 3, 256
			\end{bmatrix}\times 2 $& -   \\  \hline
			
			conv5\_x &  4$\times$4          & 
			$\begin{bmatrix}
			3\times 3, 512\\
			3\times 3, 512
			\end{bmatrix}\times 2$  & $\begin{bmatrix}
			3\times 3, 512\\
			3\times 3, 512
			\end{bmatrix}\times 2$&  $\begin{bmatrix}
			3\times 3, 512\\
			3\times 3, 512
			\end{bmatrix}\times 2 $&$\begin{bmatrix}
			3\times 3, 512\\
			3\times 3, 512
			\end{bmatrix}\times 2$  \\  \hline
			
			\multirow{2}{*}{Classifier }&  1$\times$1 &   global average pool& global average pool&global average pool&global average pool\\  \cline{2-6}
			
			& &  100D fully-connected & 100D fully-connected& 100D fully-connected & 100D fully-connected\\  \hline
			
	\end{tabular}}
	
	\label{resnet50}
\end{table}

\begin{table}
	\caption{Architectural details of WRN-16-2~\cite{zagoruyko2016wide} with auxiliary classifiers for CIFAR-100 classification.}
	\centering
	\resizebox{0.85\linewidth}{!}{
		\begin{tabular}{c|c|c|c|c}
			\hline
			Layer name & Output size & $f(\cdot)$& $b_{1}(\cdot)$& $b_{2}(\cdot)$\\ 	\hline
			conv1&  32$\times$32 &  $3\times3, 16$ &-&-\\  \hline
			conv2\_x &  32$\times$32              & 
			$\begin{bmatrix}
			3\times 3, 32\\ 
			3\times 3, 32
			\end{bmatrix}\times 2 $  &  -&  -  \\  \hline
			conv3\_x &  16$\times$16           & 
			$\begin{bmatrix}
			3\times 3, 64\\ 
			3\times 3, 64
			\end{bmatrix}\times 2 $  &  $\begin{bmatrix}
			3\times 3, 64\\ 
			3\times 3, 64
			\end{bmatrix}\times 2 $ & - \\  \hline
			conv4\_x &  8$\times$8           & 
			$\begin{bmatrix}
			3\times 3, 128\\ 
			3\times 3, 128
			\end{bmatrix}\times 2 $  &  $\begin{bmatrix}
			3\times 3, 128\\ 
			3\times 3, 128
			\end{bmatrix}\times 2 $&  $\begin{bmatrix}
			3\times 3, 128\\ 
			3\times 3, 128
			\end{bmatrix}\times 2 $\\  \hline
			
			\multirow{2}{*}{Classifier }&  1$\times$1 &   global average pool& global average pool&global average pool\\  \cline{2-5}
			
			& &  100D fully-connected & 100D fully-connected& 100D fully-connected\\  \hline
	\end{tabular}}
	
	\label{wrn162}
\end{table}

\begin{table}
	\caption{Architectural details of DenseNet-40-12~\cite{huang2017densely} with auxiliary classifiers for CIFAR-100 classification.}
	\centering
	\resizebox{1\linewidth}{!}{
		\begin{tabular}{c|c|c|c|c}
			\hline
			Layers & Output size & $f(\cdot)$& $b_{1}(\cdot)$& $b_{2}(\cdot)$\\ 	\hline
			Convolution&  32$\times$32 &  $3\times3, 16$ &-&-\\  \hline
			Dense Block (1) &  32$\times$32              & 
			$\begin{bmatrix}
			1\times 1\ \rm{conv}\\ 
			3\times 3\ \rm{conv}
			\end{bmatrix}\times 12 $  &  -&  -  \\  \hline

			\multirow{2}{*}{Transition Layer (1)}&  32$\times$32 &  $1\times 1\ \rm{conv}$ &-&-\\  \cline{2-5}
			&  16$\times$16 &  $2\times 2$ average pool, stride 2 &-&-\\  \hline

			Dense Block (2) &  16$\times$16           & 
			$\begin{bmatrix}
			1\times 1\ \rm{conv}\\ 
			3\times 3\ \rm{conv}
			\end{bmatrix}\times 12  $  &  $\begin{bmatrix}
			1\times 1\ \rm{conv}\\ 
			3\times 3\ \rm{conv}
			\end{bmatrix}\times 12  $ & - \\  \hline
			
			\multirow{2}{*}{Transition Layer (2)}&  16$\times$16 &  $1\times 1\ \rm{conv}$ &$1\times 1\ \rm{conv}$&-\\  \cline{2-5}
			&  8$\times$8 &  $2\times 2$ average pool, stride 2 &$2\times 2$ average pool, stride 2&-\\  \hline

			Dense Block (3) &  8$\times$8&$\begin{bmatrix}
			1\times 1\ \rm{conv}\\ 
			3\times 3\ \rm{conv}
			\end{bmatrix}\times 12  $  &  $\begin{bmatrix}
			1\times 1\ \rm{conv}\\ 
			3\times 3\ \rm{conv}
			\end{bmatrix}\times 12  $ &  $\begin{bmatrix}
			1\times 1\ \rm{conv}\\ 
			3\times 3\ \rm{conv}
			\end{bmatrix}\times 12  $ \\  \hline
			
			\multirow{2}{*}{Classification Layer }&  1$\times$1 &   global average pool& global average pool&global average pool\\  \cline{2-5}
			
			& &  100D fully-connected & 100D fully-connected& 100D fully-connected\\  \hline
	\end{tabular}}
	
	\label{densenet}
\end{table}

\begin{table}[tbp]
	\centering
	\caption{Architectural details of HCGNet-A1~\cite{yang2020gated} with auxiliary classifiers for CIFAR-100 classification.} 
	
	\begin{tabular}{c|c|c|c|c}  
		
		\hline
		Stage&IR&$f(\cdot)$& $b_{1}(\cdot)$ & $b_{2}(\cdot)$ \\
		\hline
		Stem&32$\times$32&3$\times$3 Conv,24&-&-\\
		\hline
		Hybrid Block&32$\times$32&SMG$\times$8 ($k=12$)&-&-\\
		Transition&32$\times$32&SMG$\times$1&SMG$\times$1&-\\
		Hybrid Block&16$\times$16&SMG$\times$8 ($k=24$)&SMG$\times$8 ($k=24$)&-\\
		Transition&16$\times$16&SMG$\times$1&SMG$\times$1&SMG$\times$1\\
		Hybrid Block&8$\times$8&SMG$\times$8 ($k=36$)&SMG$\times$8 ($k=36$)&SMG$\times$8 ($k=36$)\\
		\hline
		\multirow{2}{*}{Classification}&1$\times$1&global average pool&global average pool&global average pool\\
		&-&100D FC, softmax&100D FC, softmax&100D FC, softmax\\
		\hline
	\end{tabular}
	\label{hcgnet}
\end{table}

\begin{table}
	\caption{Architectural details of ResNet-18~\cite{he2016deep} with auxiliary classifiers for fined-grained classification. Here, $N$ denotes the number of classes.}
	\centering
	\resizebox{1\linewidth}{!}{
		\begin{tabular}{c|c|c|c|c|c}
			\hline
			Layer name & Output size & $f(\cdot)$& $b_{1}(\cdot)$& $b_{2}(\cdot)$& $b_{3}(\cdot)$\\ 	\hline
			conv1&  112$\times$112 &  $7\times7, 64$, stride 2 &-&-&-\\  \hline
			\multirow{2}{*}{conv2\_x} &  \multirow{2}{*}{56$\times$56}              & 
			3$\times$3, max pool, stride 2  &  -&  - &  -  \\  \cline{3-6}
			&&$\begin{bmatrix}
			3\times 3, 64\\
			3\times 3, 64
			\end{bmatrix}\times 2 $&  -&  - &  -\\  \hline

			conv3\_x &  28$\times$28           & 
			$\begin{bmatrix}
			3\times 3, 128\\
			3\times 3, 128
			\end{bmatrix}\times 2 $  &  $\begin{bmatrix}
			3\times 3, 128\\
			3\times 3, 128
			\end{bmatrix}\times 2 $ & - & - \\  \hline
			
			conv4\_x &  14$\times$14           & 
			$\begin{bmatrix}
			3\times 3, 256\\
			3\times 3, 256
			\end{bmatrix}\times 2 $  &  $\begin{bmatrix}
			3\times 3, 256\\
			3\times 3, 256
			\end{bmatrix}\times 2 $&  $\begin{bmatrix}
			3\times 3, 256\\
			3\times 3, 256
			\end{bmatrix}\times 2 $& -  \\  \hline
			
			conv5\_x &  7$\times$7         & 
			$\begin{bmatrix}
			3\times 3, 512\\
			3\times 3, 512
			\end{bmatrix}\times 2$  & $\begin{bmatrix}
			3\times 3, 512\\
			3\times 3, 512
			\end{bmatrix}\times 2$&  $\begin{bmatrix}
			3\times 3, 512\\
			3\times 3, 512
			\end{bmatrix}\times 2 $& $\begin{bmatrix}
			3\times 3, 512\\
			3\times 3, 512
			\end{bmatrix}\times 2 $  \\  \hline

			\multirow{2}{*}{Classifier }&  1$\times$1 &   global average pool& global average pool&global average pool&global average pool\\  \cline{2-6}
			
			& &  $N$-D fully-connected & $N$-D fully-connected& $N$-D fully-connected & $N$-D fully-connected\\  \hline
	\end{tabular}}
	
	\label{resnet18_imagenet}
\end{table}

\begin{table}
	\caption{Architectural details of ResNet-50~\cite{he2016deep} with auxiliary classifiers for ImageNet classification.}
	\centering
	\resizebox{1\linewidth}{!}{
		\begin{tabular}{c|c|c|c|c|c}
			\hline
			Layer name & Output size & $f(\cdot)$& $b_{1}(\cdot)$& $b_{2}(\cdot)$& $b_{3}(\cdot)$\\ 	\hline
			conv1&  112$\times$112 &  $7\times7, 64$, stride 2 &-&-&-\\  \hline
			\multirow{2}{*}{conv2\_x} &  \multirow{2}{*}{56$\times$56}              & 
			3$\times$3, max pool, stride 2  &  -&  - &  -  \\  \cline{3-6}
			&&$\begin{bmatrix}
			1\times 1, 64\\
			3\times 3, 64\\
			1\times 1, 256
			\end{bmatrix}\times 3 $&  -&  - &  -\\  \hline

			conv3\_x &  28$\times$28           & 
			$\begin{bmatrix}
			1\times 1, 128\\
			3\times 3, 128\\
			1\times 1, 512
			\end{bmatrix}\times 4 $  &  $\begin{bmatrix}
			1\times 1, 128\\
			3\times 3, 128\\
			1\times 1, 512
			\end{bmatrix}\times 4 $ & - & - \\  \hline
			
			conv4\_x &  14$\times$14           & 
			$\begin{bmatrix}
			1\times 1, 256\\
			3\times 3, 256\\
			1\times 1, 1024
			\end{bmatrix}\times 6 $  &  $\begin{bmatrix}
			1\times 1, 256\\
			3\times 3, 256\\
			1\times 1, 1024
			\end{bmatrix}\times 6 $&  $\begin{bmatrix}
			1\times 1, 256\\
			3\times 3, 256\\
			1\times 1, 1024
			\end{bmatrix}\times 6 $& -  \\  \hline
			
			conv5\_x &  7$\times$7         & 
			$\begin{bmatrix}
			1\times 1, 512\\
			3\times 3, 512\\
			1\times 1, 2048
			\end{bmatrix}\times 3$  & $\begin{bmatrix}
			1\times 1, 512\\
			3\times 3, 512\\
			1\times 1, 2048
			\end{bmatrix}\times 3$&  $\begin{bmatrix}
			1\times 1, 512\\
			3\times 3, 512\\
			1\times 1, 2048
			\end{bmatrix}\times 3 $& $\begin{bmatrix}
			1\times 1, 512\\
			3\times 3, 512\\
			1\times 1, 2048
			\end{bmatrix}\times 3 $  \\  \hline

			\multirow{2}{*}{Classifier }&  1$\times$1 &   global average pool& global average pool&global average pool&global average pool\\  \cline{2-6}
			
			& &  1000D fully-connected & 1000D fully-connected& 1000D fully-connected & 1000D fully-connected\\  \hline
	\end{tabular}}
	
	\label{resnet50_imagenet}
\end{table}
\clearpage
\section{Experimental setup}
\subsection{Image Classification}  
\subsubsection{Dataset}
\begin{itemize}
	\item \textbf{CIFAR-100}~\cite{krizhevsky2009learning} is a
	standard image classification dataset, containing 50k training images and 10k
	test images in 100 classes.
	\item \textbf{CUB-200-2011}~\cite{wah2011caltech} contains 200 species of birds
	with 5994 training images and 5794 test images.
	\item \textbf{Standford Dogs}~\cite{khosla2011novel} contains
	120 breeds of dogs with 12000 training images and 8580 test images.
	\item \textbf{MIT67}~\cite{quattoni2009recognizing} contains 67 indoor categories with 5356 training images and 1337 test images.
	\item \textbf{Stanford Cars}~\cite{krause20133d} contains 196 classes of cars with 8144 training images and 8041 testing images.
	\item \textbf{FGVC-Aircraft}~\cite{maji2013fine} contains 100 classes of aircraft variants with 6667 training images and 3333 testing images.
	\item \textbf{ImageNet}~\cite{deng2009imagenet} is a large-scale image classification dataset, which contains 1.28
	million training images and 50k validation images in 1000 classes. ImageNet is also a hierarchical dataset that includes both coarse- and fine-grained class distinction.
\end{itemize}
\textbf{Data pre-processing}. We utilize the standard data pre-processing pipeline~\cite{huang2017densely}, \emph{i.e.} random cropping and flipping. The resolution of each input image is 32$\times$32 in CIFAR-100 and 224$\times$224 in fine-grained datasets and ImageNet.

\subsubsection{Training details} 
\begin{itemize}
	\item \textbf{CIFAR-100:} all network are trained by stochastic gradient descent (SGD) optimizer with a momentum of 0.9, a weight decay of $5\times 10^{-4}$, and a batch size of 128. We start at 5 epochs for linear warm-up from 0 to an initial learning rate of 0.1, which avoids the possible model collapse issue for data augmentation and Self-KD training. Then the learning rate is divided by 10 after the 105-th and 155-th epochs within the total 205 epochs.
	\item \textbf{Fine-grained classification:} all network are trained by stochastic gradient descent (SGD) optimizer with a momentum of 0.9, a weight decay of $1\times 10^{-4}$, and a batch size of 32. We start at 5 epochs for linear warm-up from 0 to an initial learning rate of 0.1, which avoids the possible model collapse issue for data augmentation and Self-KD training. Then the learning rate is divided by 10 after the 105-th and 155-th epochs within the total 205 epochs.
	\item \textbf{ImageNet:} all network are trained by stochastic gradient descent (SGD) optimizer with a momentum of 0.9, a weight decay of $1\times 10^{-4}$, and a batch size of 256. We start at 5 epochs for linear warm-up from 0 to an initial learning rate of 0.1, which avoids the possible model collapse issue for data augmentation and Self-KD training. Then we use a cosine learning rate scheduler from an initial learning rate of 0.1 to 0 throughout the 300 epochs.
	
\end{itemize}

\subsubsection{Object Detection}
\begin{itemize}
	\item \textbf{COCO-2017}~\cite{lin2014microsoft} contains 120k training images and 5k validation images. In this paper, we adopt 5k validation images for test.
\end{itemize}

\textbf{Data pre-processing}. We utilize the default data pre-processing of MMDetection~\cite{chen2019mmdetection}. The shorter side of the input image is resized to 800 pixels, the longer side is limited up to
1333 pixels.

\textbf{Training details}. We adopt a 1x training schedule with a momentum of 0.9 and a weight decay of 0.0001. We start at 500 linear warm-up iterations from 0 to an initial learning rate of 0.02. Then the learning rate is divided by 10 after the 8-th and 11-th epochs within the total 12 epochs. Training is conducted on 8 GPUs using synchronized SGD with a batch
size of 1 per GPU. 

\subsubsection{Semantic Segmentation}
\begin{itemize}
	\item \textbf{Pascal VOC}~\cite{everingham2010pascal} contains 10582/1449/1456 images for train/val/test with 21 semantic categories. Some training images are augmented with extra annotations provided by Hariharan \emph{et al.} \cite{hariharan2011semantic}.
	\item \textbf{ADE20K}~\cite{zhou2017scene} contains 20k/2k/3k images for train/val/test with 150 semantic categories.
	\item \textbf{COCO-Stuff-164k}~\cite{caesar2018coco} covers 172 labels and contains 164k images: 118k for training, 5k for validation, 20k for test-dev and 20k for the test-challenge.
\end{itemize}
\textbf{Data pre-processing}. In this paper's semantic segmentation experiments, we retain the original training images and use validation images for test. Following the standard data augmentation~\cite{yang2022cross}, we employ random flipping and scaling in the range of [0.5, 2]. During the training phase, we use a crop size of $512\times 512$. During the test phase, we utilize the original image size.

\textbf{Training details}. All experiments are optimized by SGD with a momentum of 0.9, a batch size of 16 and an initial learning rate of 0.02. The number of the total training iterations is 40K. The learning rate is decayed by $(1-\frac{iter}{total\_iter})^{0.9}$ following the polynomial annealing policy~\cite{chen2017rethinking}. Training is conducted on 8 GPUs using synchronized SGD with a batch size of 2 per GPU. The implementation is based on an open codebase\footnote{https://github.com/winycg/CIRKD} released by Yang \emph{et al.}~\cite{yang2022cross}.

\end{document}